\documentclass[conference]{IEEEtran}
\IEEEoverridecommandlockouts

\usepackage{cite}
\usepackage{amsmath,amssymb,amsfonts}
\usepackage{algorithmic}
\usepackage{graphicx}
\usepackage{textcomp}
\usepackage{xcolor}
\usepackage{url}
\usepackage{enumitem}
\usepackage{booktabs}

\def\BibTeX{{\rm B\kern-.05em{\sc i\kern-.025em b}\kern-.08em
    T\kern-.1667em\lower.7ex\hbox{E}\kern-.125emX}}
    
\begin{document}

\title{Leveraging Retrieval-Augmented Generation for Culturally Inclusive Hakka Chatbots: Design Insights and User Perceptions}

\author{\IEEEauthorblockN{Chen-Chi Chang}
\IEEEauthorblockA{
\textit{National United University}\\
Taiwan \\
kiwi@gm.nuu.edu.tw}
\and
\IEEEauthorblockN{Han-Pi Chang}
\IEEEauthorblockA{
\textit{National Central University}\\
Taiwan \\
hanbi@cc.ncu.edu.tw}
\and
\IEEEauthorblockN{Hung-Shin Lee}
\IEEEauthorblockA{\textit{United Link Co., Ltd.} \\
Taiwan \\
hungshin.lee@united-link.com.tw}
}

\maketitle
\begin{abstract}
In an era where cultural preservation is increasingly intertwined with technological innovation, this study introduces a groundbreaking approach to promoting and safeguarding the rich heritage of Taiwanese Hakka culture through the development of a Retrieval-Augmented Generation (RAG)-enhanced chatbot. Traditional large language models (LLMs), while powerful, often fall short in delivering accurate and contextually rich responses, particularly in culturally specific domains. By integrating external databases with generative AI models, RAG technology bridges this gap, empowering chatbots to not only provide precise answers but also resonate deeply with the cultural nuances that are crucial for authentic interactions. This study delves into the intricate process of augmenting the chatbot's knowledge base with targeted cultural data, specifically curated to reflect the unique aspects of Hakka traditions, language, and practices. Through dynamic information retrieval, the RAG-enhanced chatbot becomes a versatile tool capable of handling complex inquiries that demand an in-depth understanding of Hakka cultural context. This is particularly significant in an age where digital platforms often dilute cultural identities, making the role of culturally aware AI systems more critical than ever. System usability studies conducted as part of our research reveal a marked improvement in both user satisfaction and engagement, highlighting the chatbot's effectiveness in fostering a deeper connection with Hakka culture. The feedback underscores the potential of RAG technology to not only enhance user experience but also to serve as a vital instrument in the broader mission of ethnic mainstreaming and cultural celebration. This paper demonstrates the potential of RAG technology in crafting culturally aware conversational AI systems that contribute to ethnic mainstreaming and the celebration of cultural diversity.
\end{abstract}
\section{Introduction}
\label{sec:intro}

In the era of globalization, societies worldwide are increasingly characterized by their cultural diversity, with multiple ethnic groups coexisting within a single national framework. This multicultural landscape presents challenges, particularly in preserving the unique identities of minority cultures while promoting their integration into the broader societal context. Among these, the Hakka culture in Taiwan stands out as a poignant example of a rich, yet often underrepresented, ethnic heritage seeking its rightful place in the digital age \cite{wang2007diaspora}. The importance of cultural preservation and promotion in such multi-ethnic societies cannot be overstated \cite{thao2023preserving}, as it not only safeguards the heritage of minority groups but also enriches the social fabric by fostering mutual understanding and respect among diverse populations. Some tools, such as LLaMA, ChatGPT, and Gemini, has become a groundbreaking instrument in Arts and Humanities in the rapidly evolving technological landscape, enabling novel engagement between humans and artificial intelligence (AI) \cite{Rane2023}. 

However, previous AI models, particularly Large Language Models (LLMs), have shown limitations in adequately capturing and conveying the nuances of specific cultural narratives \cite{colas2022language}. This is primarily due to their reliance on parameterized knowledge, which, while effective for general inquiries, falls short in contexts requiring deep, culturally specific insights. The introduction of Retrieval-Augmented Generation (RAG) technology represents a significant leap forward, promising to enhance the accuracy, reliability, and cultural relevance of AI-driven chatbots by leveraging external databases to supplement the chatbot's responses with accurate and contextually appropriate information \cite{Gao2023}.

This study merges LLMs with RAG to develop culturally inclusive Hakka chatbots \cite{jeong2023generative}. These advanced models integrate the comprehensive learning capabilities of pre-trained sequence-to-sequence models with a specialized, non-parametric memory system, consisting of a dense vector index filled with Hakka cultural knowledge and dictionaries. By leveraging a neural retriever, these chatbots can access and utilize specific cultural insights, addressing the limitations of LLMs in handling knowledge-intensive tasks and providing more accurate, culturally aware interactions. This innovative approach enhances the chatbots' performance on tasks requiring deep cultural understanding and pioneers a path towards solving the challenges of decision provenance and knowledge updating in AI systems \cite{Lewis2020}. This study aims to fill these gaps by comprehensively analysing a RAG-enhanced Hakka narrative chatbot's design, implementation, and user perceptions. By doing so, it seeks to contribute to the broader discourse on leveraging advanced AI technologies for cultural preservation and representation, offering new insights into the potential of RAG to bridge the divide between technological innovation and the intricate demands of cultural inclusivity.

\section{AI Bias Against Minority Groups}

Incorporating LLMs into societal applications poses ethical challenges, notably the risk of exacerbating biases in training data. This issue is particularly critical for minorities, as it can lead to biased outcomes and reinforce stereotypes, undermining fairness and causing harm\cite{liyanage2023ethical}. Sampling bias arises when a dataset disproportionately favours certain instances over others, making it unrepresentative of the broader reality. This form of bias is prevalent across various data collections and significantly impacts the performance and generalization of algorithms \cite{Srinivasan2021}. For instance, in low-resource languages and minority cultures, datasets might predominantly include languages and cultural references that are widely recognized or have a larger presence in digital and media landscapes.

Consequently, languages and cultures with fewer resources or digital visibility are underrepresented. This underrepresentation can lead to inadequate performance of algorithms in processing, understanding, or generating content related to these low-resource languages and minority cultures. As a result, the learned algorithms may struggle to accurately recognize or produce linguistic and cultural nuances specific to these groups, thereby exacerbating the digital divide and cultural invisibility. Thus, addressing sampling bias is crucial for developing more inclusive and equitable technologies that accurately reflect and serve the diversity of global languages and cultures. Previous research underscores the complex interplay between AI and race, revealing how AI contributes to unequal opportunities for individuals from certain racial backgrounds while also possessing the ability to uncover racial discrimination \cite{Intahchomphoo2020}. The issue of AI bias is particularly detrimental to minority ethnic groups, as it amplifies existing disparities and hampers the fair detection and remediation of racial biases. To address the challenge of biased AI and its adverse effects on minority groups, a multifaceted approach is necessary. This involves enriching training datasets with a broad spectrum of racial representations, developing algorithms to identify and amend biases, and cultivating a development ethos emphasizing ethical practices and inclusivity. By adopting these measures, the goal is to advance AI technologies that are just and reflective of our diverse global community.

\section{Strategies for Retrieval of Minority Languages and Cultures}

This study explores three strategies to enhance the retrieval of accurate information on minority group languages and cultures by LLMs within conversational AI robots \cite{jiang2021language}. Initially, the method of utilizing precisely designed prompt words combined with internet resource retrieval is proposed \cite{hassan2023chatgpt}. This approach, leveraging search engines based on accurate and comprehensive keywords, allows for the holistic acquisition of knowledge about minority group languages and cultures. Secondly, the feasibility of applying RAG is discussed. By establishing a dedicated knowledge base for minority group languages and cultures and training it with RAG technology, the system prioritizes fetching data from this knowledge base when queries related to minority group languages and cultures are made. The results are then fed back to the LLM, which subsequently delivers the response to the user in a natural language conversational manner. Thirdly, the paper suggests the continuous collection and organization of language and cultural materials of minority groups. Through training and establishing a specialized LLM, users can directly interact using an LLM tailored to minority group languages and cultures. These three modalities not only aid in improving the precision and quality of responses by LLMs regarding minority group languages and cultures but also provide technical support for the protection and promotion of the heritage of minority group languages and cultures. Implementing these methods effectively fosters the protection of linguistic and cultural diversity while also enhancing the visibility and accessibility of minority group languages and cultures. This offers a new technological pathway for promoting cultural diversity and inclusivity.

\section{Precisely Designed Prompts}

The prior research has highlighted the critical role of precise, prompt engineering in enhancing the retrieval capabilities of LLMs across diverse fields, including entrepreneurship, art, science, and healthcare \cite{mesko2023prompt}. By meticulously designing prompts, it is possible to access more obscure information, circumventing the pitfalls of AI illusions that lead to inaccuracies \cite{chen2023development}. This tailored approach to prompt construction ensures that LLMs can produce results that are more precise and profoundly aligned with the inquiry's specific context, thereby mitigating errors associated with AI's interpretative limitations \cite{Henrickson2023}. Incorporating precisely designed prompts combined with Google Web Search into conversational AI chatbots represents a pivotal strategy for enhancing the accurate retrieval of information related to minority languages and cultures. By tailoring prompts to include specific keywords and phrases deeply embedded within the unique contexts of minority cultures, AI systems can leverage Google's vast index of web content to fetch more relevant and culturally nuanced information. This approach broadens the scope of data the AI can draw upon and improves the precision of search results by aligning them more closely with the user's query intent. Such integration facilitates a more informed and context-aware retrieval process, ensuring that chatbots' responses are accurate and resonate with the cultural depth and linguistic subtleties of minority groups. Consequently, this method significantly contributes to overcoming the limitations of traditional AI models, which often struggle with the complexity and diversity inherent in minority languages and cultures, thereby promoting a more inclusive digital landscape.

\subsection{Retrieval-Augmented Generation}

Open Domain Question Answering (ODQA) serves as a pivotal challenge in the field of Natural Language Processing (NLP), dedicated to generating responses to factual inquiries based on extensive knowledge corpora, without relying on explicitly provided evidence \cite{Zhang2022}. It plays a crucial role in advancing natural language understanding. Among recent innovations in ODQA \cite{quarteroni2009designing}, RAG stands out for enhancing question-answering capabilities by integrating with external, domain-specific knowledge bases \cite{Siriwardhana2023}. This technology holds particular promise for amplifying the representation and accessibility of minority cultures within ODQA systems. By leveraging RAG, these systems can significantly improve in delivering precise and contextually relevant information of diverse cultural backgrounds, thereby supporting the broader goal of inclusivity in digital knowledge spaces. Establishing a dedicated knowledge base for minority languages/cultures and training it with RAG technology marks a major advancement in improving the accurate retrieval of minority languages/cultures in AI chatbots. This approach involves compiling a comprehensive repository of linguistic data, cultural nuances, historical contexts, and societal values specific to minority groups. AI chatbots can dynamically retrieve and incorporate this specialized information during conversations by integrating this knowledge base with RAG technology. The RAG framework enhances the chatbot's ability to understand and generate responses deeply rooted in the cultural and linguistic context of the inquiry. This method ensures the generated responses are contextually accurate, culturally respectful, and informative. Training AI chatbots with such a rich, curated knowledge base significantly mitigates the risk of perpetuating stereotypes or overlooking the complexities of minority cultures. As a result, this strategy elevates the user experience by providing more meaningful and relevant interactions. It plays a crucial role in preserving and promoting the richness of minority languages and cultures within the digital domain.

\subsection{LLMs for Minority Group Cultures}

Pre-trained multilingual language models have demonstrated remarkable efficacy in handling tasks across multiple languages, significantly enhancing the application of NLP for languages with limited resources \cite{hangya2022improving}. Despite these advancements, certain languages, including Hakka, continue to pose challenges, with current multilingual models struggling to achieve optimal performance \cite{Yang2022}.

Utilizing machine translation for low-resource to mainstream languages in constructing LLMs risks erasing cultural nuances and omitting minority ethnic groups' culture and knowledge, posing a significant threat to cultural preservation \cite{ranathunga2023neural}. Establishing LLMs specifically for minority group cultures represents a transformative approach to improving the accurate retrieval of minority languages and cultures in AI chatbots. By developing LLMs trained exclusively on linguistic patterns, cultural references, and historical contexts unique to minority groups, these models become adept at processing and understanding queries related to these specific cultures with high accuracy and sensitivity. This specialized training ensures that the AI chatbots are well-equipped to handle the nuances and complexities inherent in minority languages and cultures, enabling them to generate contextually relevant and culturally informed responses. Such an approach addresses the common challenge of generic LLMs failing to adequately represent the depth and diversity of minority cultures due to a lack of targeted data. By prioritizing the development of culture-specific LLMs, AI technologies can offer more inclusive and respectful interactions, facilitating a digital environment where minority languages and cultures are accurately represented and celebrated. This, in turn, enhances the user experience for individuals seeking information or engagement with these cultures, fostering a greater understanding and appreciation of cultural diversity.

\begin{figure}[t]
\centering
\includegraphics[width=\linewidth]{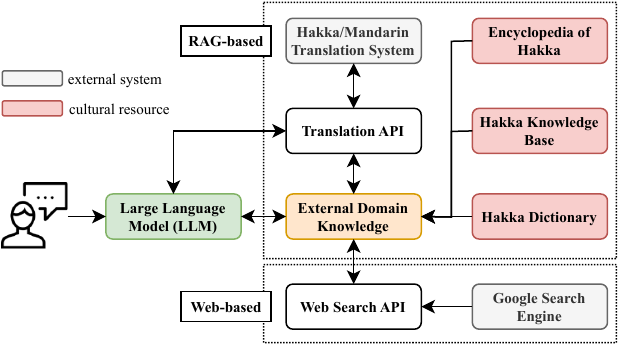}
\vspace{-15pt}
\caption{Our proposed system architecture.}
\label{fig:architecture}
\vspace{-15pt}
\end{figure}

\section{Research Methodology}

\subsection{System Architecture}

As seen in Figure \ref{fig:architecture}, this study integrated the precisely designed prompts with Google Web Search Engine (\url{google.com}) and RAG within the Coze platform (\url{coze.com}) to enhance the accurate retrieval of information on minority languages and cultures in AI chatbots. The system architecture is centralized around a LLM that processes user queries. For translation, it directs inquiries to a Hakka/Mandarin Translation System via a Translation API. For cultural questions, it taps into trained External Domain Knowledge. Queries outside these domains are resolved using a Web Search API linked to Google Search Engine, with the findings integrated by the LLM into a coherent response.

\subsection{RAG for Minority Languages and Cultures}
 
Implementing RAG to dynamically utilize a dedicated knowledge base for minority group languages and cultures within the Coze platform represents a strategic enhancement for conversational AI chatbots. This involves creating a comprehensive and curated knowledge base with resources like the Hakka dictionary, the Ministry of Education's Hakka Knowledge Database, and the Hakka Cultural Encyclopedia. The RAG technology leverages this dedicated knowledge base to provide contextually relevant and culturally nuanced responses to queries related to Hakka language and culture. When a query is received, the RAG system consults this specialized knowledge base to find the most relevant information before generating a response. This process ensures that the chatbot's replies are accurate and deeply reflective of Hakka cultural heritage. By integrating RAG with tools such as Hakka translation system for language translation, Coze can offer a conversational AI that is exceptionally adept at handling inquiries specific to the Hakka minority group, thereby fostering greater understanding and appreciation of this rich cultural heritage. 

\vspace{5pt}
\noindent
i. \textbf{Prompts Design for RAG}:

\vspace{5pt}
\noindent
\fbox{
\begin{minipage}{0.95\linewidth}
\setlength{\parskip}{0.2em}
\textbf{\#\# Role}: You will be conversing with an expert in the Hakka language and culture, proficient and willing to communicate in Traditional Chinese.

\textbf{\#\# Skill 1 (Inquiries into Taiwanese Hakka Culture)}:

This expert can provide answers using the ``Knowledge'' resources, including the Hakka dictionary, the Ministry of Education's Hakka Knowledge Database, the Hakka Cultural Encyclopedia, Hakka Characteristic Words, and key Hakka towns and townships, answering questions primarily within the Taiwanese Hakka context.

\textbf{\#\# Skill 2 (Mandarin to Hakka Translation)}:

When a user inputs Hakka or is preparing to translate Mandarin into Hakka, the system is called to execute the translation process. The Hakka translation system is utilized to convert entire Mandarin queries into Hakka.

\textbf{\#\# Limitations}
\begin{enumerate}[left=0pt, itemsep=0.5em, parsep=0pt]
\item Answers are provided solely based on the user's questions. No prompts are provided.
\item Data is returned following the aforementioned format.
\item Tasks are limited to those related to the Hakka dictionary, the Ministry of Education's Hakka Knowledge Database, Hakka Cultural Encyclopedia, Hakka Characteristic Words, and key Hakka towns and townships.
\item The expert specializes in language translation, particularly from Mandarin to Hakka. If your question falls outside of this scope, an appropriate response may not be provided.
\item Responses should be in Hakka characters, not Romanized phonetics. No additional reference information is to be provided. The expert's translation knowledge is limited to data provided by ``HakkaTrans'' and ``Knowledge''.
\end{enumerate}
\end{minipage}
}

\vspace{10pt}
\noindent
ii. \textbf{Establishing a Comprehensive Knowledge Base for RAG Implementation in Hakka Cultural Studies}:

A meticulously curated knowledge base has been established to enable the RAG method within the Coze platform to enhance conversational AI's capability to accurately retrieve and generate responses related to Hakka culture and language. This knowledge base comprises several pivotal components designed to encapsulate the breadth and depth of Hakka cultural heritage and linguistic nuances: 1) \textbf{Encyclopedia of Hakka Cultural Affairs.} This extensive resource provides a wealth of information on the history, traditions, and cultural practices unique to the Hakka community, serving as a foundational element for generating informed responses. 2) \textbf{Hakka Culture Knowledge Base of the Ministry of Education.} Developed by the Ministry of Education, this database offers authoritative insights into Hakka culture, educational content, and scholarly articles, enriching the chatbot's ability to provide accurate cultural explanations. 3) \textbf{Hakka Dictionary.} A comprehensive lexicon that includes Hakka vocabulary definitions, pronunciations, and usage examples. This dictionary is instrumental in ensuring the linguistic accuracy of the chatbot's responses in both Hakka and Mandarin. 4) \textbf{Hakka Characteristic Words.} A curated list of words uniquely significant to Hakka language and culture. This list aids the chatbot in recognizing and emphasizing culturally resonant terms, enhancing the relevance of its interactions. 5) \textbf{List of Key Hakka Towns and Villages.} Detailed information about geographic locations that are historically and culturally significant to the Hakka people. This component helps the chatbot to contextualize inquiries related to specific regions, offering insights into the local customs, dialects, and cultural landmarks. Together, these elements form a robust knowledge base that underpins the RAG method's success on the Coze platform. By integrating these resources, the AI chatbots are equipped to navigate the complexities of Hakka culture and language, offering users a rich, interactive experience that is both educational and engaging. This knowledge base not only facilitates the generation of accurate and contextually relevant responses but also plays a crucial role in preserving and promoting the rich heritage of the Hakka community through digital technology.

\section{Results and Evaluation}

Analyzing the demographic data collected from 02/29/2024 to 03/03/2024 through a survey on the Taiwanese Hakka Culture AI Chatbot usage, alongside the System Usability Scale (SUS) score \cite{bangor2008empirical}, provides a comprehensive understanding of the user base and the chatbot's current performance metrics. The demographic data is shown in Table 1. The assessment of the prototype system underwent a two-phase process. During the first phase, the SUS yielded a score of 61. Subsequently, in the second phase, adjustments were made to the prototype system based on user feedback gathered during the initial phase. These modifications included the incorporation of chat in Hakka features, as well as translation functions. Following these enhancements, the SUS score rose to 69, exceeding 68 for a more universally positive user experience \cite{lewis2018system}.

\begin{table}[t]
\centering
\caption{Demographic data with respect to genders, ages, and education backgrounds and SUS Scores.}
\vspace{0pt}
\label{tab:my_label}
\begin{tabular}{cccc}
\toprule
Male & 27 & Female & 44\\
\midrule
Under 18 years old & 16 & 19-30 years old & 3\\
31-40 years old & 4 & 41-50 years old & 18\\
51-60 years old & 15 & & \\
\midrule
Below high school & 22 & College/University & 27\\
Master's degree & 18 & PhD & 4\\
\midrule
\midrule
Phase I SUS Score & 61.81 & Phase II SUS Score & 69.52\\
\bottomrule
\end{tabular}
\vspace{-5pt}
\end{table}

\section{Discussion}

The contribution of this article lies in its revelation that when employing RAG for the development of chatbots tailored for minority cultures, merely expanding the ethnic-cultural knowledge base isn't sufficient. It underscores the importance of integrating a translation system capable of facilitating seamless communication between minority languages and mainstream languages. This integration enables the chatbot to communicate in minority languages, enhancing inclusivity and accessibility within multicultural contexts.

\subsection{Challenges Identified}

Users encountered several challenges interacting with the chatbot, including unsuccessful responses to specific cultural/linguistic queries, limitations in understanding/translating Hakka characters, and delayed response time. Notably, the chatbot struggled with translating phrases related to cultural festivities and explaining unique Hakka terms, reflecting a gap in its knowledge base and linguistic processing capabilities. The challenges highlight the need for expansive and detailed linguistic databases encompassing the breadth of cultural expressions and dialectal variations within Hakka. Furthermore, the issues with translation accuracy and character recognition emphasize the necessity for sophisticated NLP techniques that can navigate the complexities of minority languages.

\subsection{Recommendations for Improvement}

In response to user feedback, a comprehensive strategy has been proposed to enhance the chatbot's effectiveness, particularly in representing Hakka culture. The proposed measures include augmenting the knowledge base with a wide range of Hakka cultural resources to improve response precision significantly. Furthermore, implementing advanced NLP algorithms is recommended to navigate contextual nuances, idiomatic expressions, and dialectal variations, refining the chatbot's translation and interpretation capabilities. Optimizing the user interface is advocated to ensure smoother navigation and interaction, thereby boosting user engagement and satisfaction. Additionally, integrating adaptive learning mechanisms through machine learning algorithms could enable the chatbot to evolve its response strategies based on user interactions, addressing response relevance and accuracy challenges. Lastly, machine translation features are proposed to facilitate communication between low-resource and mainstream languages, offering users a more accessible interaction modality. These recommendations aim to address identified shortcomings and improve the overall user experience with the chatbot, showcasing a targeted approach toward developing more sophisticated, user-friendly, and culturally inclusive AI tools.

\section{Conclusions}

The enhancement of prompt design coupled with the integration of Google Web Search offers a limited improvement in the accuracy of retrieved information. Additionally, the dynamic nature of online resources presents challenges in maintaining consistent quality, indicating that this approach warrants further observation. Establishing LLMs for minority languages and cultures encounters the significant hurdle of resource scarcity. Relying solely on machine translation to convert mainstream language texts into those of minority groups, followed by training, risks diluting cultural meanings and the essence of actual cultural content, potentially exacerbating data bias. Currently, the most viable solution appears to be adopting the RAG technique for the AI conversational technologies of minority languages. This method not only leverages the NLP capabilities of LLMs but also preserves the cultural connotations inherent in minority languages and cultures. By integrating RAG, AI technologies can balance linguistic proficiency and cultural integrity, offering a promising avenue for developing culturally inclusive AI dialog systems. 

This approach underscores the importance of nuanced language and cultural understanding in creating AI technologies that are both sophisticated and sensitive to the diverse tapestry of human cultures. These strategies hold significant academic and practical implications for understanding and enhancing the application of AI technologies against the backdrop of global multiculturalism, offering broader perspectives and deeper insights into the development of AI technologies.

\bibliographystyle{IEEEtran}
\bibliography{references}

\end{document}